\documentclass{article}

    \PassOptionsToPackage{numbers, compress}{natbib}

 \usepackage[dblblindworkshop, final]{neurips_2025}
\workshoptitle{Workshop on Space in Vision, Language, and Embodied AI}

\usepackage[utf8]{inputenc} 
\usepackage[T1]{fontenc}    
\usepackage{hyperref}       
\usepackage{url}            
\usepackage{booktabs}       
\usepackage{amsfonts}       
\usepackage{nicefrac}       
\usepackage{microtype}      
\usepackage[dvipsnames]{xcolor}
\usepackage{graphicx} 
\usepackage{amsmath}
\usepackage{subcaption} 

\title{DenseScan: Advancing 3D Scene Understanding \\
with 2D Dense Annotation}

\author{%
Zirui Wang$^1$ \quad Tao Zhang$^2$ \\
$^1$University of Illinois at Urbana-Champaign \quad $^2$Wuhan University \\
\texttt{ziruiw3@illinois.edu} \quad \texttt{zhang\_tao@whu.edu.cn}
}

\begin{document}
\maketitle
\begin{abstract}
3D understanding is a key capability for real-world AI assistance. High-quality data plays an important role in driving the development of the 3D understanding community.
Current 3D scene understanding datasets often provide geometric and instance-level information, yet they lack the rich semantic annotations necessary for nuanced visual-language tasks.
In this work, we introduce DenseScan, a novel dataset with detailed multi-level descriptions generated by an automated pipeline leveraging multi-view 2D images and multimodal large language models (MLLMs). Our approach enables dense captioning of scene elements, ensuring comprehensive object-level descriptions that capture context-sensitive details. Furthermore, we extend these annotations through scenario-based question generation, producing high-level queries that integrate object properties, spatial relationships, and scene context. By coupling geometric detail with semantic richness, DenseScan broadens the range of downstream tasks, from detailed visual-language navigation to interactive question answering. Experimental results demonstrate that our method significantly enhances object-level understanding and question-answering performance in 3D environments compared to traditional annotation pipelines. We release both the annotated dataset and our annotation pipeline to facilitate future research and applications in robotics, augmented reality, and beyond. Through DenseScan, we aim to catalyze new avenues in 3D scene understanding, allowing researchers and practitioners to tackle the complexities of real-world environments with richer, more contextually aware annotations.
\end{abstract}    
\section{Introduction}
\label{sec:introduction}

Understanding 3D world is a crucial in various applications such as robotics and autonomous driving, where agents are expected to know about the surrounding environment and carry out complex tasks based on human's instructions. However, current 3D MLLMs \cite{3dllm,zhu2024llava, zhu2024empowering, SegPoint} perform poorly in these practical applications and there are still significant gaps between their capabilities and the requirements of real-world applications. The scarcity of high-quality 3D multimodal data severely limits the development of 3D MLLMs.
Although there are many multimodal datasets~\cite{chen2020scanrefer, achlioptas2020referit3d, reason3d, zhu2024empowering} in the current 3D community available for training 3D MLLMs, these datasets such as ScanRefer~\cite{chen2020scanrefer}, ReferIt3D~\cite{achlioptas2020referit3d}, and Reason3D~\cite{reason3d} only focus on localizing 3D objects based on short, direct or implicit text referring. Besides, recent advances in 3D Question Answering datasets \cite{ma2022sqa3d, huang2023embodied,azuma_2022_CVPR} emphasize integrating point clouds, images, and textual data to form rich multimodal representations, but LLM-based QA-pairs generation often introduce hallucinated information and contextual
misalignment. As a result, they cannot effectively inject rich 3D knowledge into models during training, nor can they effectively benchmark how far current MLLMs are from real-world applications. Unlike prior referring setups optimized for short disambiguation, our scenario-driven segmentation requires models to parse multi-sentence context, inter-object relations beyond the target, and functional cues to produce precise masks.

In the past, 3D multimodal data relied on manual annotation, which was extremely costly and made it difficult to produce a rich diversity and massive quantity of annotations to drive rapid community development. Some recent works~\cite{reason3d, zhu2024empowering, SegPoint} have attempted to design automated pipelines to generate 3D multimodal data, such as using GPT-like models to refine existing textual descriptions rather than generating new ones from direct scene analysis. These models, though effective in producing fluent text, lack a deep understanding of spatial relationships, object affordances, and occlusions, as they operate in a predominantly text-driven space without strong visual grounding.


Currently, 2D MLLMs such as GPT-4o~\cite{achiam2023gpt}, Gemini, Qwen2.5 VL~\cite{yang2024qwen2}, and InternVL 2.5~\cite{chen2024far} demonstrate powerful capabilities, including fine-grained understanding of visual signals and robust reasoning abilities. The responses generated by these models even far exceed the quality of data produced by most skilled human annotators. However, the quality of 3D multimodal datasets is significantly lower than that in the 2D community, and the intelligence level of 3D MLLMs lags far behind 2D MLLMs.



In this paper, we aim to design a systematic and efficient pipeline that leverages 2D MLLMs to produce multi-level (including object-level and scene-level) high-quality 3D data, in order to further advance the 3D MLLM community. Specificly, we propose an automated pipeline to annotate ScanNet \cite{dai2017scannet} with detailed scene-level descriptions. 

Our automated annotation system is organized into four distinct stages: 1) we initiate the process with multi-level object captioning, where state-of-the-art 2D MLLMs generate detailed captions for objects in multi-view 2D images, effectively capturing rich semantic information; 2) These initial captions undergo a rigorous phase of filtering and consistency checking, along with style adaptation, to transform them into natural-flow referring expressions that are coherent and contextually appropriate; 3) leverages multi-level object captions, we generate scenario-driven questions by prompting the MLLM annotator to draw on its world knowledge to suggest multiple frequent events or interactions that are typical for similar scenes. It is important to note that these event cues are not necessarily accurate depictions of what is happening in the specific scene, but rather informed hypotheses based on general knowledge; they are then seamlessly combined with object descriptions to form complex scenario-driven referring expressions that reflect intricate contextual relationships; 4) the generated annotations are subjected to LLM verification, augmented by human-assisted filtering, to ensure their accuracy and establish robust validation benchmarks. The resulting dataset, DenseScan, comprises 1,513 scenes and 20,113 object instances from ScanNet \cite{dai2017scannet}, featuring 38,765 dense referring expressions and 37,483 scenario-based questions. Compared to existing referring datasets, DenseScan stands out by offering significantly richer linguistic diversity and more precise object annotations while maintaining a comparable scale.

Along With the DenseScan dataset, we introduce a novel task called scenario-driven segmentation, which requires models to locate objects and generate segmentation masks based on detailed, long-text descriptions. Compared to traditional referring segmentation, which typically involves short referring expressions tied to specific objects, scenario-driven segmentation presents a greater challenge by incorporating complex contextual cues, spatial relationships, and functional attributes within a scene. This task pushes models to move beyond simple object identification and short referring expression learning, requiring deeper semantic understanding and reasoning about object interactions within diverse environments. 

To tackle the challenge of complex scenario-driven reasoning in 3D point clouds, we propose Dense3D, a novel LLM-based framework that uniquely embeds multi-modal information rather than relying solely on 2D image inputs. Our model consists of three key components: a Point Encoder, which applies voxelization and a Sparse 3D U-Net backbone to extract point-wise features, further refined through a superpoint pooling layer for computational efficiency; Multi-Modal LLMs, which integrate both 2D multi-view images and 3D point cloud representations (from depth map and camera paramters) to enhance spatial and semantic understanding; and a 3D Query Decoder, a transformer-based module that translates high-level textual cues into precise segmentation masks. By fusing 3D point clouds, 2D multi-views images, and textual descriptions—Dense3D achieves deeper contextual reasoning and more precise segmentation, setting a new standard for multi-modal 3D scene understanding.

In summary, our contributions are threefold:
\begin{enumerate}
    \item \textbf{Automated Annotation System.} We have developed an automated annotation system that leverages state-of-the-art 2D MLLMs to generate reliable and high-quality 3D multimodal data. Based on this pipeline, we have created the DenseScan dataset, which features rich, multi-level annotations—including detailed object-level captions and comprehensive scene-level descriptions—that better capture complex spatial relationships and contextual cues.
    \item \textbf{Scenario-Driven Segmentation.} We have manually reviewed and refined a more challenging benchmark that incorporates a novel task: \emph{scenario-driven segmentation}, with requires models to generate precise segmentation masks guided by long, context-rich descriptions, pushing the boundaries of traditional referring segmentation and better reflecting real-world application demands.
    \item \textbf{Dense3D Framework.} We have introduced Dense3D, a 3D MLLM framework that effectively fuses 3D point cloud data, 2D multi-view images, and textual information. Through a Point Encoder, a Multi-Modal LLMs, and a 3D Query Decoder, we achieve deeper contextual reasoning and more accurate segmentation, thus taking a step further in multi-modal 3D scene understanding.
\end{enumerate}

\section{Related Work}
\label{sec:related}

\begin{figure*}
  \centering
  \includegraphics[clip,width=0.95\linewidth]{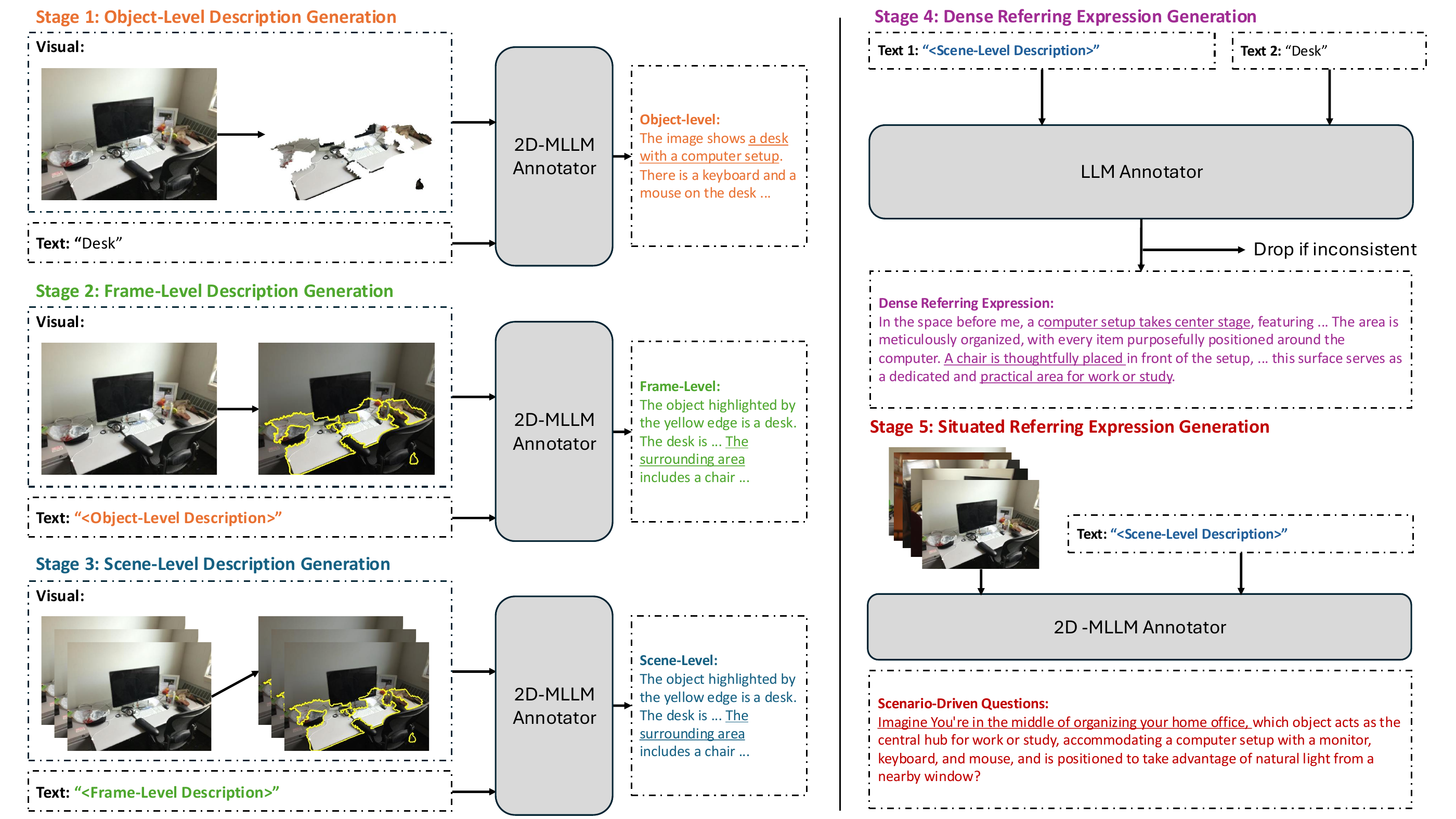}
  \caption{\noindent\textbf{Overview of DenseScan Data Generation Pipeline}. \textit{Stage 1:} crop the target object and generate object-level description; \textit{Stage 2:} highlight target object in a single frame and generate frame-level description to capture spacial dependencies with the surroundings; \textit{Stage 3:} use multiple frames with target object highlighted to compose scene-level descriptions; \textit{Stage 4:} Raw scene-level description need to go through a LLM for consistency checking, and in-consistent description will be eliminated; \textit{Stage 5:} Adopt MLLM annotator to generate scenario-driven questions and verified by LLM and human before release to benchmark.}
  \label{pipeline_overview}
\end{figure*}

\noindent\textbf{Datasets for 3D Scene Understanding}. Advances in 3D computer vision are deeply intertwined with the availability of large-scale, high-quality datasets\cite{Silberman:ECCV12, armeni20163d, chang2017matterport3d,behley2019semantickitti,hackel2017semantic3d}. In the realm of 3D instance segmentation, large-scale real-world datasets are limited. Early benchmarks like ScanNet\cite{dai2017scannet} and S3DIS\cite{armeni20163d} provided a solid foundation by offering detailed scans of real-world indoor environments captured using RGB-D cameras or Matterport systems. Subsequently, researchers have increasingly built upon these public datasets by incorporating diverse, richly detailed annotations \cite{chen2020scanrefer, achlioptas2020referit3d,zhang2023multi3drefer,reason3d,jiang2024multimodal3dreasoningsegmentation,zhu2024empowering}. Notably,  ScanRefer\cite{chen2020scanrefer} provides a set of natural language referring expressions for objects in indoor 3D scenes. Subsequent efforts, such as ReferIt3D\cite{achlioptas2020referit3d}, refines the annotation process through more fine-grained object categorization and the inclusion of multiple object instances per scene, and Multi3DRefer\cite{zhang2023multi3drefer}, expands the task scope by accommodating descriptions that reference zero, one, or multiple objects, which better mirrors the complexities of real-world scenarios. More recently, several efforts have pushed the boundary even further by designing implicit object descriptions that require higher-level reasoning, such as Reason3D\cite{reason3d}, ScanReason\cite{zhu2024empowering}, Instruct3D\cite{SegPoint} and ReasonSeg3D\cite{jiang2024multimodal3dreasoningsegmentation}. Despite these advancements, our work develop a denser and longer text description for objects, capturing detailed semantic attributes, spatial context, and subtle visual cues that are often overlooked in earlier datasets. Besides, while previous reasoning-based questions tend to be abstract and focus on broad, often decontextualized logic or inference, scenario-like questions are tightly grounded in the spatial and semantic context of a specific scene. Together, the new dataset provide a more comprehensive framework for advancing 3D scene understanding and segmentation. 


\noindent\textbf{3D Multi-Modal Large Language Models}. Inspired by the powerful reasoning abilities of Large Language Models, researchers have injected LLMs into vision domain\cite{alayrac2022flamingo,li2023blip, lai2024lisa,zhu2023minigpt, liu2023visual}, namely multi-modal large language models (MLLMs). Upon prior works \cite{li2023blip, alayrac2022flamingo} at integrating visual context with language models, recent efforts has been made to develop MLLMs that seamlessly integrating diverse capabilities for instruction-based tasks. For instance, VisionLLM\cite{wang2023visionllm} offers a vision-centric interface through instruction tuning, though it doesn’t support advanced reasoning. Another novel line of research has emerged with the debut of LISA\cite{lai2023lisa}, alongside several subsequent studies \cite{rasheed2024glamm, ren2024pixellm}that have significantly advanced the field of multimodal large language models (MLLMs) in 2D space. Building on the advancements in 2D multimodal language models, researchers are now venturing into the realm of 3D MLLMs, aiming to enhance spatial understanding and unlock novel applications in complex volumetric environments. 3D-LLM\cite{hong20233d} leverage 2D foundation models to inject 3D spatial understanding into language models. PointLLM\cite{xu2024pointllm} build on top of LLaVA \cite{liu2023visual} to train LLM with 3D point cloud representations. 

\noindent\textbf{Grounding MLLMs}. Recent work has aimed to fully leverage the reasoning capabilities of large language models for addressing 3D downstream tasks such as segmentation and localization. In particular, SegPoint\cite{SegPoint} and Reason3D\cite{reason3d} focus on interpreting complex textual descriptions of individual objects, whereas ReasonSeg3D\cite{jiang2024multimodal3dreasoningsegmentation} targets multi-object referring expression segmentation with accompanying text explanations. ScanReason\cite{zhu2024empowering} developed comprehensive and hierarchical 3D reasoning grounding benchmark that assess fundamental-understanding of 3D world to high-level reasoning skills. Our work distinguishes itself from previous efforts by incorporating denser text descriptions extracted from detailed 2D visuals, while our model leverages both point cloud and 2D visual information to enhance reasoning capabilities.

\section{DenseScan}
\label{sec:dataset}



Existing 3D indoor scene datasets \cite{chen2020scanrefer, achlioptas2020referit3d, zhang2023multi3drefer} typically rely on brief, one-sentence annotations or fragmented short-phrase labels that fail to capture the full semantic and spatial complexity of diverse environments. Moreover, traditional annotation pipelines often overlook the rich contextual details available from 2D RGB-D frames, thereby missing the fine-grained visual cues essential for robust scene understanding. Although more recent datasets \cite{SegPoint, reason3d, zhu2024empowering} employ large language models such as GPT-4 \cite{achiam2023gpt} to generate diverse referring expressions, they still struggle to incorporate the inherent visual context of 2D spaces. ReasonSeg3D \cite{jiang2024multimodal3dreasoningsegmentation} leverages GPT4o—which supports visual input—by providing a single scene image with its ground-truth segmentation to enhance 3D spatial comprehension; however, this approach is constrained by the reliance on reconstructed 3D scene images that do not fully reflect the true attributes of the objects. DenseScan addresses these deficiencies by offering extended, detailed text descriptions that comprehensively depict each scene and by integrating an advanced generation pipeline that harnesses intricate details from 2D scans. This enhanced approach not only enriches the descriptive quality of the dataset but encourage more accurate and nuanced 3D scene analysis.

\begin{figure*}[ht]
  \centering

  \begin{subfigure}[t]{0.59\textwidth}
    \centering
    \includegraphics[clip, trim={0.5cm 2cm 0.6cm 0.3cm}, width=\linewidth]{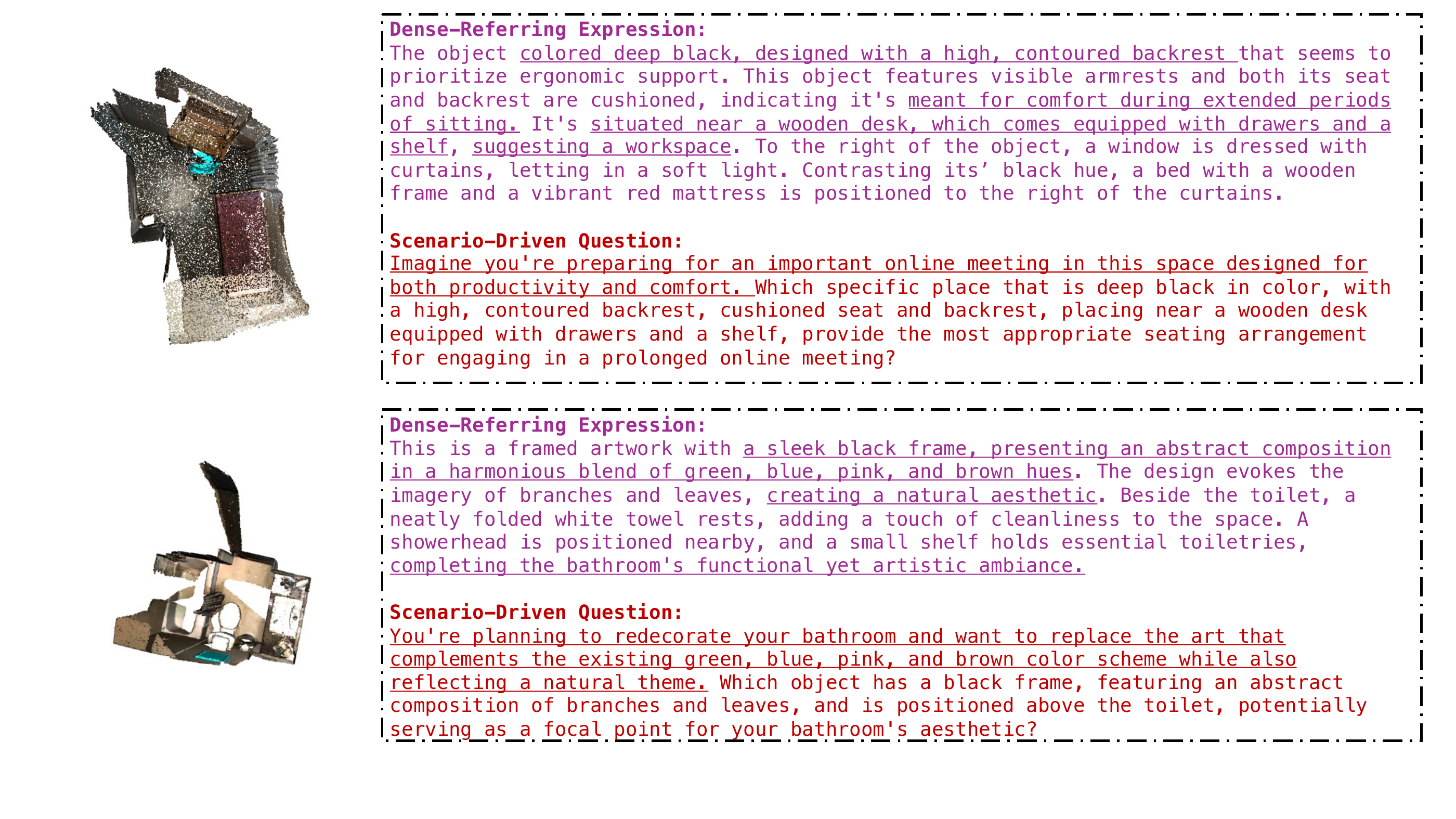}
    \label{fig:example_data_sub}
  \end{subfigure}
  \hfill
  \begin{subfigure}[t]{0.38\textwidth}
    \centering
    \includegraphics[clip, width=\linewidth]{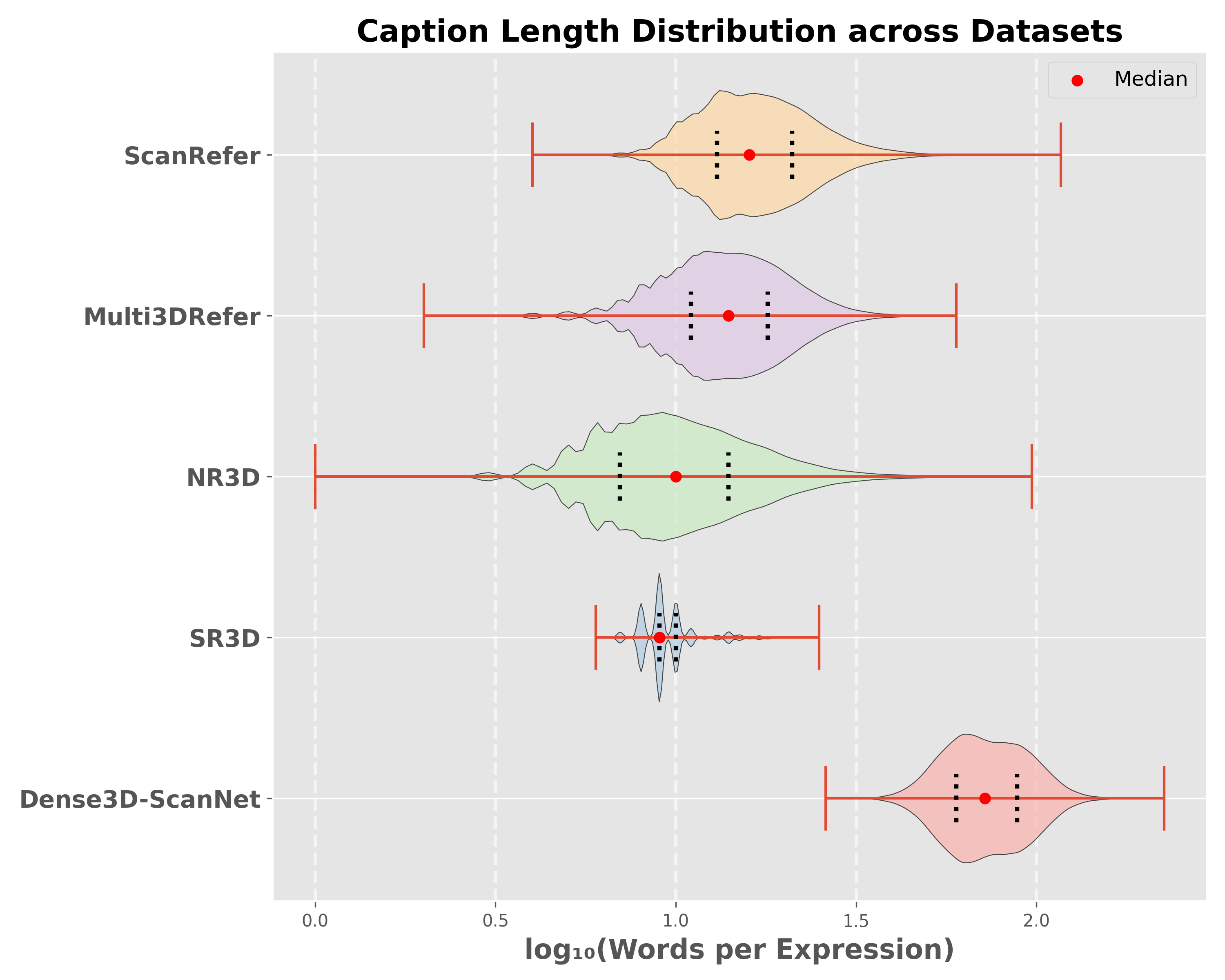}
    \label{fig:caption_length_stats_sub}
  \end{subfigure}

  \caption{\textbf{DenseScan Dataset.} (a) Two objects sampled from ScanNet\cite{dai2017scannet} with dense-referring expressions and scenario-driven questions. Objects shown are \textcolor{Cyan}{"office chair"} and \textcolor{Cyan}{"pictures"}, both highlighted in the point cloud. (b)  The distribution of description lengths, with dotted lines marking quartiles. The x-axis is scaled logarithmically (base-10) to handle the long-tail distribution.}
  \label{fig:combined_dense_scan}
\end{figure*}

DenseScan build upon a widely used 3D scene understanding dataset ScanNet \cite{dai2017scannet} by directly using the multi-view 2D RGB-D video frames along with its semantic annotations for rich context extraction. Specifically, DenseScan enrich each 3D scans with dense object-level captions and scenario-specific questions, providing a richer context for downstream tasks such as referring expression segmentation, visual grounding and question answering. An example sample of DenseScan and the detailed statistics and distribution of the dataset is shown in Table \ref{tab:dataset_stats} and Figure \ref{fig:combined_dense_scan}, respectively. 

\subsection{Data Annotation Pipeline}
We carefully designed an automatic annotation pipeline to generate dense scene-level descriptions and scenario-based referring object text expressions for ScanNet dataset \cite{dai2017scannet}, as illustrated in Fig. \ref{pipeline_overview}. The pipeline leverages the InterVL2-76B \cite{chen2024far} multimodal large language model (MLLM) as the multimodal data annotator and Qwen2 \cite{yang2024qwen2} as our LLM assistant for consistency checking,filtering out unidentified objects and conflicting object descriptions. 


\noindent\textbf{Object-Level Description}. To obtain the object-level dense caption rich is instance-specific semantics information, we first select the single video frame with the largest object area from the video frames and mask out non-object pixels. In this way, object isolation minimizes the interference from the background or nearby objects and directs the MLLM annotator to focus solely on annotated object's features. The cropped object image is then fed into InterVL2-76B to generate detailed descriptions. 

\noindent\textbf{Frame-Level Description}. Building upon the detailed single-object descriptions, we enrich the single-object text description by incorporating spatial and relational information from the selected video frame. Specifically, we enhance the visual prominence of the annotated object using yellow contours, which serves as a visual cue to isolate the object from its surroundings while preserving the contextual detials of the scene. The annotated image is fed into the MLLM annotator along with a carefully designed prompt. The prompt is tailored to instruct the model to focus on generating a caption that not only describes the highlighted object but also captures its spatial relationship and interactions with adjacent elements. 

\noindent\textbf{Scene-Level Description}. To acquire a comprehensive description of the scene, we uniformly sampled 8 frames from the video, ensuring that diverse perspectives is captured. We also apply a yellow contour to consistently highlight the target objects. These frames, along with frame-level descriptions, are processed by InterVL2-76B though a text prompt to generate scene-level object captions. The scene-level caption is further processed by an LLM annotator to transform into natural-flow referring expressions that are coherent, named \emph{dense referring expression}.

\noindent\textbf{Scenario-Like Questions}. In the final stage of our pipeline, we aim to generate scenario-like questions that target the unique aspects of the selected object. Using the scene-level caption generated from target object, combing with a comprehensive list of all other objects present in the 3D environment, we carefully prompt the LLM annotator to craft scenario-like questions that specifically refer only to the target object. These questions are designed to evoke contextual reasoning and encourage nuanced interpretations that highlight the role, functionality, and significance of the object within realistic scene scenarios. 


\noindent\textbf{Benchmark Quality Control}. Before releasing scenario-driven questions for benchmakring, we apply additional quality checks to the existing scenario-driven questions. Specifically, we provided these questions to a LLM and prompt it to identify the referring objects-along with unique characteristics for the objects to verify that the description accurately represents the object. Any descriptions that are inconsistent are eliminated. Additionally, following \cite{chen2020scanrefer}, we manually eliminate under-sampled objects such as "stick", and structural objects such as "ceiling", "wall" to keep a balanced distribution of commonly referenced objects. Finally, we perform manual reviews on a random subset of the remaining descriptions to ensure overall quality and consistency.

\subsection{Dataset Statistics}
We quantitatively compare DenseScan with existing 3D scene benchmarks along two key dimensions: data scale and referring expression length.

\noindent\textbf{Data Scale}. DenseScan comprises 1,513 scenes and 20,113 object instances in total from ScanNet \cite{dai2017scannet}, with 38,765 dense-referring expressions and 37,483 scenario-based questions. Following ScanNet \cite{dai2017scannet}, we partition it into a training set of 1201 scans and a validation set of 312 scans. As can be seen from Table \ref{tab:dataset_stats}, DenseScan provides comparable number of referring expressions than existing datasets like ScanRefer \cite{chen2020scanrefer}, Multi3DRefer \cite{zhang2023multi3drefer}. This is achieved through the special design of automated data collection pipeline with MLLM annotator, which is more robust and time-efficient compared to labor-intensive labeling.

\begin{table*}
\centering
\begin{tabular}{lccccc}
\toprule
 & \textbf{\# of Scenes} &  \textbf{\# of Descriptions} &\textbf{Annotation Method} \\
\midrule
ScanRefer\cite{chen2020scanrefer} & 703 &  51,583 & human labeling\\
Sr3D\cite{achlioptas2020referit3d} & 707 & 83,572 & human labeling\\
Nr3D\cite{achlioptas2020referit3d} & 707 & 41,503 & human labeling\\
Multi3DRefer\cite{zhang2023multi3drefer} & 800 & 61,926 & human + ChatGPT\\
Reason3D\cite{reason3d} & - & 2,484 & GPT-4\cite{achiam2023gpt}\\
Instruct3D\cite{SegPoint} & 280 & 2,565  & - \\
ScanReason\cite{zhu2024empowering} & 1,456 & 12,929 & GPT-4\cite{achiam2023gpt}\\ 
ReasonSeg3D\cite{jiang2024multimodal3dreasoningsegmentation} & 1,513 & 20,113 & GPT-4V\cite{2023GPT4VisionSC}\\
\textbf{DenseScan (Ours)} & \textbf{1,513} & \textbf{76,248} & InterVL2.5\cite{chen2024far} + Qwen2\cite{yang2024qwen2}\\
\bottomrule
\end{tabular}
\caption{\noindent\textbf{Comparison of various 3D referring expression datasets}. showing the number of scenes, number of descriptions, and annotation methods used. The proposed DenseScan features 1,513 scenes, 76,248 descriptions (including 38,765 dense-referring expressions and 37,483 scenario-based questions), and an automated annotation pipeline, surpassing previous datasets in scale.}
\label{tab:dataset_stats}
\end{table*}

\noindent\textbf{Expression Length}. Caption length is another important feature of DenseScan is its emphasis on generating detailed, instance-specific captions. In our dataset, each object is described with referring expressions that are notably longer and more descriptive than those in other benchmarks. While traditional datasets often offer brief captions that focus on basic attributes, DenseScan delivers captions that, on average, contain a higher word count—reflecting complex semantic details and contextual relationships. This extended caption length facilitates deeper semantic parsing, enabling models to capture intricate object properties and inter-object relationships that are critical for high-level reasoning in 3D scenes. 

\subsection{Evaluation Metrics}
Following traditional 3D segmentation methods \cite{huang2021text, wu20243dgres, SegPoint,kolodiazhnyi2024oneformer3d,schult2023mask3d}, we adopt mean intersection over union (mIoU), which quantifies the overall alignment between the predicted and ground-truth point cloud by averaging the Intersection over Union scores over all 3D point clouds. We also use Accuracy (Acc) to measures the percentage of segmentation predictions that achieve an IoU above a specified threshold $k = \{0.25, 0.5\}$

\section{3D Scenario-Driven Segmentation}
\label{sec:method}

\begin{figure*}
  \centering
  \includegraphics[clip,trim={0.5cm 6cm 0.6cm 2cm}, width=0.95\linewidth]{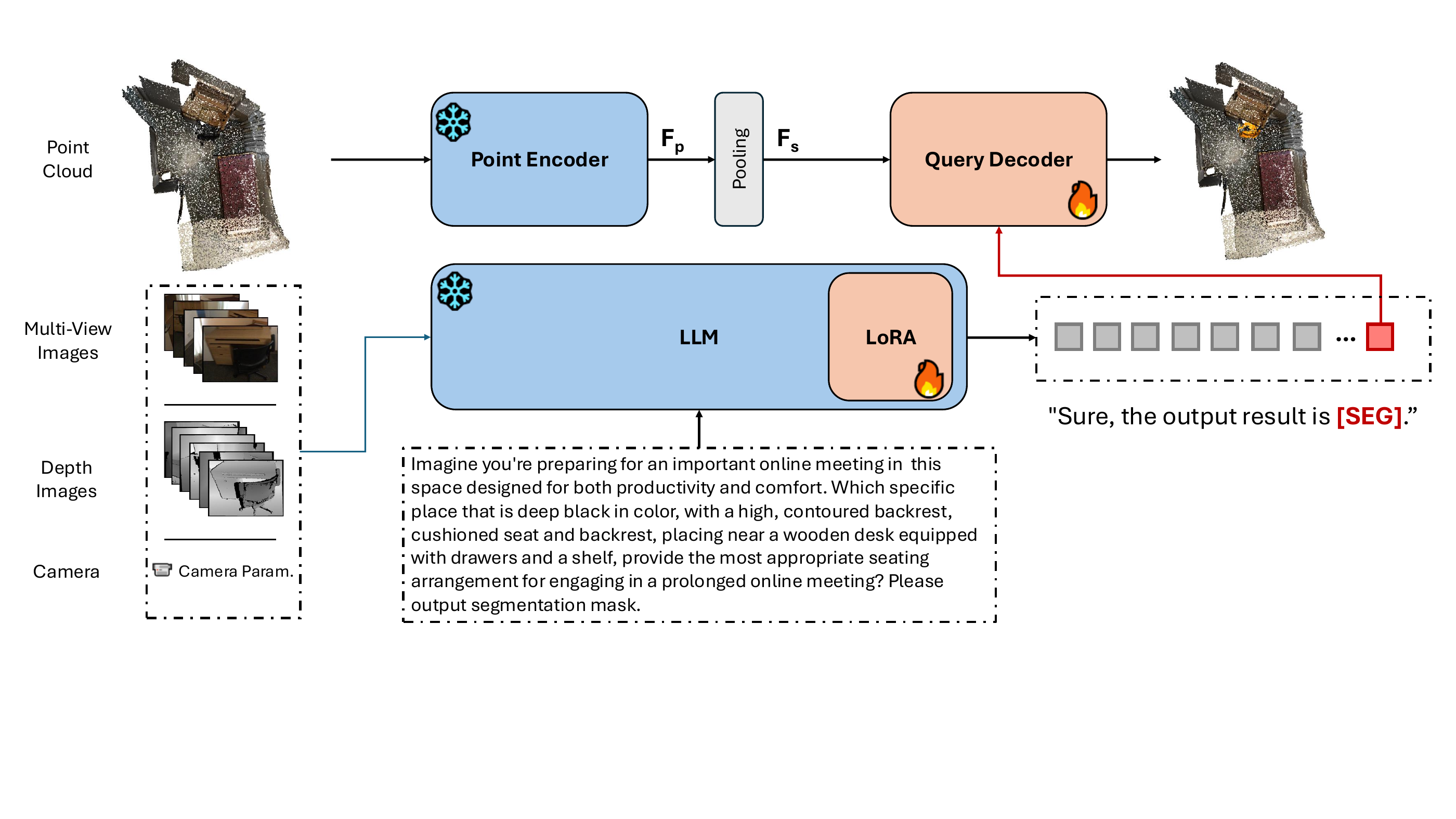}
  \caption{\noindent\textbf{Dense3D Model Architecture}. Given the 3D scene and language description, the model first reprocess 3D scene into multimodal 2D data, including RGB frames, depth map and camera poses. Depth map and camera poses composes the 3D positional embedding, along with the RGB frames and text descriptions to be send into the LLM. Output from LLM include special \textsc{\textcolor{Red}{[SEG]}} token that is crucial to guide the Query Decoder for mask generation.  }
  \label{baseline}
\end{figure*}

\subsection{Task Definition}
Scenario-driven segmentation extends traditional segmentation by incorporating rich, scenario-specific semantic cues, capturing the intended real-world scenario depicted by the annotations. Specifically, 3D scenario-driven segmentation task involves generating a 3D segmentation map $\mathcal{M}$ from a given point cloud representation $\mathcal{P}$ of a scene, along with a long-context scenario-like question $\mathcal{X}_{txt}$, extending beyond traditional referring expression segmentation tasks with short text descriptions \cite{huang2021text, wu20243dgres}. \\

\subsection{Vanilla Baseline}
We design a simple LLM-based framework for this task to for complex scenario-driven reasoning within 3D point clouds. The overall architecture of Dense3D is illustrated in Figure \ref{baseline}.

\noindent\textbf{Point Encoder}. We first apply a novelization operation to the raw point cloud to discretize the spacial data. A Sparse 3D U-Net backbone \cite{graham20183d} is employed to extract point-wise features, represented as $F_p \in \mathbb{R}^{N_p \times C_p}$, where $N$ is the number of sampled points and $C$ denotes the feature dimension. To mitigate computational complexity, these features are further processed through a superpoint pooling layer that leverages pre-computed superpoints \cite{landrieu2018large}. Aggregating point-wise features using average pooling results in superpoint features $F_s \in \mathbb{R}^{N_s \times C_p}$, where $N_s$ represents the number of superpoints. 

\noindent\textbf{Multi-Modal LLMs}. To fuse 2D information in to 3D point cloud to allow model to perform better understanding capabilities, we instruct LLM to learn detail semantics leverage the strong 2D understanding priors of 2D MLLMs. We adopt pre-trained LLaVA-like models as our 2D multi-modal LLM backbone, which comprises a visual encoder, a visual projection layer, and an LLM. The visual encoder processes 2D multi-view videos of scanned 3D scenes enriched with depth and pose information—to extract robust visual features. These features are then mapped into visual tokens through the projection layer. The visual tokens, in conjunction with the corresponding text tokens, are subsequently fed into the LLM to generate textual predictions. 

To enable prediction of mask features, we follow prior works \cite{lai2023lisa, SegPoint, reason3d, jiang2024multimodal3dreasoningsegmentation} and expand LLM's vocabulary with a special \textsc{[SEG]} token, indicating a request for a 3D segmentation mask.

\noindent\textbf{3D Query Decoder}. In the final stage, we route the \textsc{[SEG]} token into a decoder module which built on transformer-based architecture to decode a segmentation mask directly from the superpoint features $F_s$. By integrating the semantic directive provided by \textsc{[SEG]} with the detailed point-level features, the decoder effectively translates high-level textual cues into precise segmentation outputs for the target 3D scene regions.

\noindent\textbf{Training Objectives}. Our model jointly optimize both the multi-modal language generation and the 3D segmentation capabilities. The overall loss is formulated as a weighted combination of the three components as follows:
$$
\begin{aligned}
    \mathcal{L} = \lambda_{LLM}\mathcal{L}_{LLM} + \lambda_{BCE}\mathcal{L}_{BCE} + \lambda_{DICE}\mathcal{L}_{DICE}
\end{aligned}
$$
where $\mathcal{L}_{LLM}$ ensure coherent and accurate text generation, and $\mathcal{L}_{BCE}$ and $\mathcal{L}_{DICE}$ help refine high-quality segmentation masks. 
\subsection{Experiments}
\label{sec:experiments}
\noindent\textbf{Instruction Tuning Datasets}. Our training dataset contains the following three types of datasets: 1) semantic segementation datasets include ScanNet200 \cite{rozenberszki2022language}; 2) referring expression datasets includes ScanRefer \cite{chen2020scanrefer}, Multi3DRefer\cite{zhang2023multi3drefer} and ReferIt3D \cite{achlioptas2020referit3d} as short-text referring expression datasets and DenseScan as long-text referring expression dataset; 3) question answering dataset include ScanQA.

\noindent\textbf{Implementation Detail}. The model is trained on 8 NVIDIA A100 GPUs. We adopt the AdamW optimizer with a learning rate of 3e-4 and use a learning rate scheduler WarmupDecay LR with the warmup steps of 100. The total batch size is set to be 16. The loss weight parameters $\lambda_{TXT}$ is set to 1.0, and weight parameters for segmentation $\lambda_{DICE}$ and $\lambda_{BCE}$ is set to 1.0 and 1.0 respectively. We adopt the pretrained point cloud encoder following Sparse 3D-Unet \cite{huang2021text} as the 3D visual encoder. We initialize our model with the weight of LLaVA-3D \cite{zhu2024llava}, and durring training, we use LoRA\cite{hu2022lora} to efficiently finetune the Large Language Model to reduce the computation costs while preserving the original 3D scene understanding capability.

\begin{table}
\centering
\begin{tabular}{lcccc}
\toprule
 & \textbf{Acc@0.25} & \textbf{Acc@0.50} & \textbf{mIOU}\\
\midrule
3D-STMN\cite{wu20243dstmn} & 20.8 & 13.2 & 14.4\\
MDIN\cite{wu20243dgres} & 21.2 & 10.8 & 15.1 \\
\textbf{Dense3D* (Ours)} & 34.3 & 20.1 & 23.2\\
\textbf{Dense3D (Ours)} & \textbf{35.3} & \textbf{20.9} & \textbf{24.0} \\
\bottomrule
\end{tabular}
\caption{Quantitative results on the 3D scenario-driven segmentation task on \textbf{DenseScan} benchmark among Dense3D (ours) and existing methods. * represents removing DenseScan dataset from the instruction tuning dataset. }
\label{tab:3dscenario_full}
\end{table}

\begin{figure*}
  \centering
  \includegraphics[clip, width=0.95\linewidth]{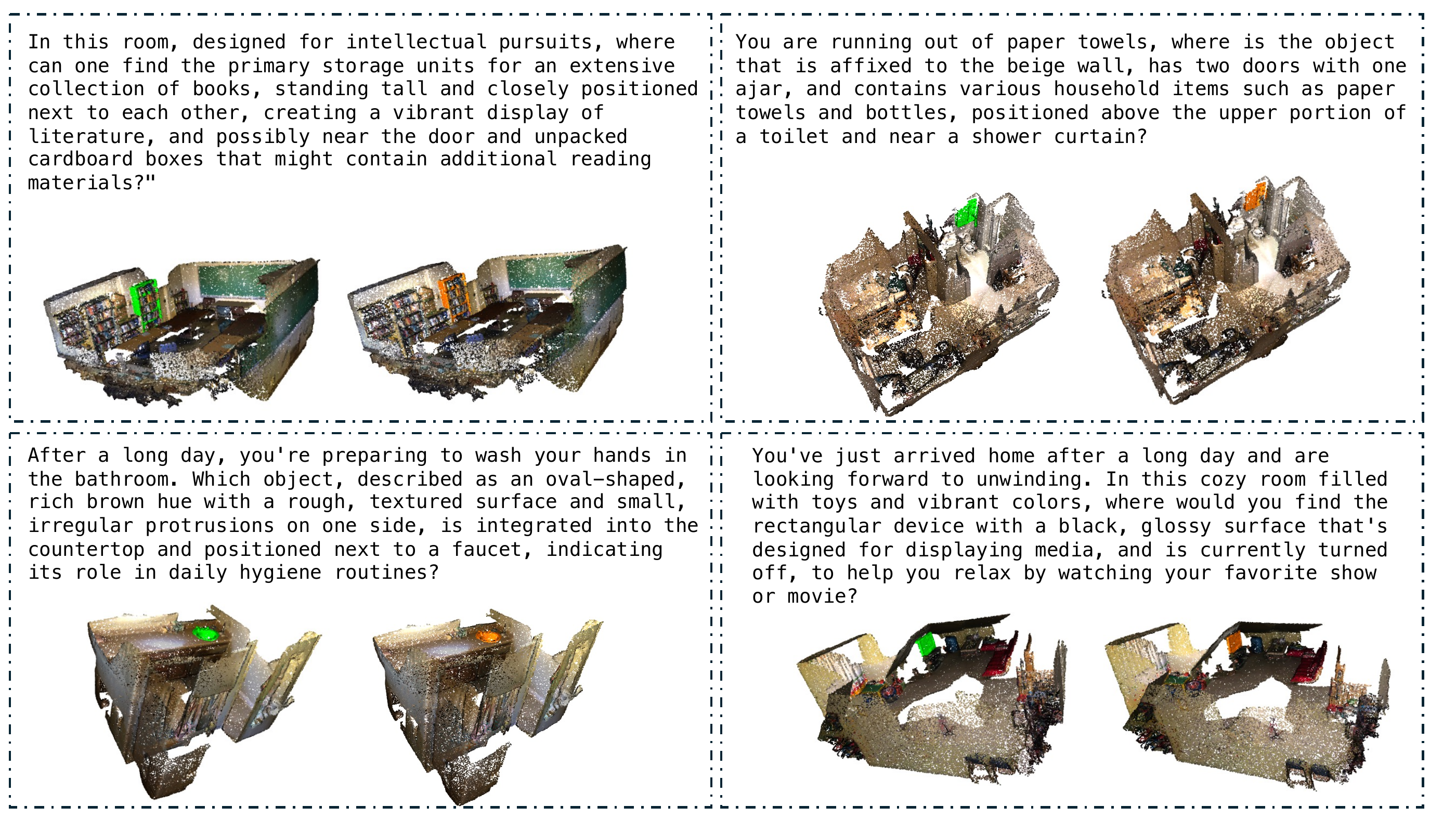}
  \caption{\noindent\textbf{Visual results for 3D Scenario-Driven Segmentation Task}. Each visual output presents a textual scenario-driven question, along with ground truth point cloud in \textcolor{green}{green} regions and predicted point cloud in \textcolor{orange}{orange} regions.}
  \label{fig:qualitative}
\end{figure*}

\noindent\textbf{Evaluation on 3D Scenario-Driven Segmentation}. The performance of 3D Scenario-Driven Segmentation is evaluated on the validation set of DenseScan, and it is shown in Table \ref{tab:3dscenario_full}. We performed a comparison between our baseline model to existing methods of various types, inncluding LLM-based methods and non-MLLM methods. Besides, to better validate the performance of our model and ensure a fair comparison, we removed DenseScan dataset from the training data, and denoted Dense3D*. From Table 1, we can see that LLM-based method generally out-perform non-LLM methods by a large amount, demonstrating its strong 3D scene understanding and reasoning capabilities over long-text referring expressions, and our baseline Dense3D achieve competitive performance on DenseScan, outperforming most existing baselines. We also show qualitative results in Figure \ref{fig:qualitative}. 

\begin{table}
\centering
\begin{tabular}{lcccc}
\toprule
 & \textbf{Acc@0.25} & \textbf{Acc@0.50} & \textbf{mIOU}\\
\midrule
TGNN\cite{huang2021text} & 38.6 & 32.7 & 28.8 \\
X-RefSeg3D\cite{qian2024x} & 40.2 & 33.5 & 30.6\\ 
3D-STMN\cite{wu20243dstmn} & 54.6 & \textbf{39.8} & 39.5 \\
SegPoint\cite{SegPoint} & - & - & \textbf{41.7} \\
\textbf{Dense3D* (Ours)} & \textbf{56.8} & 33.7 & 36.6\\
\textbf{Dense3D (Ours)} & 56.5 & 34.3 & 37.8 \\
\bottomrule
\end{tabular}
\caption{Quantitative results on the \textbf{ScanRefer} benchmark among Dense3D (ours) and exisiting methods (including specialist methods such as 3D-STMN\cite{wu20243dstmn} and LLM generalist such as SegPoint \cite{SegPoint}. * represents removing DenseScan dataset from the instruction tuning dataset.}
\label{tab:scanrefer_full}
\end{table}


\begin{table}
\centering
\begin{tabular}{lcccc}
\toprule
 & \textbf{Acc@0.25} & \textbf{Acc@0.50} & \textbf{mIOU}\\
\midrule
M3DRef-CLIP\cite{zhang2023multi3drefer} & 55.7 & 37.5 & 37.4 \\
SegPoint\cite{SegPoint} & - & - & 36.1 \\
\textbf{Dense3D* (Ours)} & 57.1 & 36.0 & 38.4\\
\textbf{Dense3D (Ours)} & \textbf{57.7} & \textbf{37.5} & \textbf{39.2} \\
\bottomrule
\end{tabular}
\caption{Quantitative results on the \textbf{Multi3DRefer} benchmark among Dense3D (ours) and existing methods. * represents removing DenseScan dataset from the instruction tuning dataset. }
\label{tab:multi3drefer_full}
\end{table}

\noindent\textbf{Does Dense Object Descriptions help 3D segmentation?}
Long-text referring expressions provide extensive details that enhance a model’s generalization by capturing nuanced attributes and maintaining long-range dependencies to differentiate similar objects. An interesting question, however, is whether dense object descriptions also benefit traditional short-context referring expressions. As shown in Table \ref{tab:scanrefer_full} and Table \ref{tab:multi3drefer_full}, training on such rich descriptions enables our model to learn fine-grained associations between language and visual features, ultimately improving its generalization. Notably, on ScanRefer\cite{chen2020scanrefer}, despite being trained on long-text data rather than optimized for short, structured expressions like specialist models (e.g., 3D-STMN), our model achieves comparable performance in identifying referents in complex real-world scenarios. This demonstrates the effectiveness of leveraging long-text multimodal training, suggesting that our approach is not only competitive with specialized methods but also more adaptable to diverse and realistic tasks.

\section{Conclusion}
\label{sec:conclusion}

In this paper, we propose a pipeline that leverages state-of-the-art 2D MLLMs to generate high-quality, multi-level 3D annotations, culminating in the DenseScan dataset, which significantly enhances linguistic diversity and contextual richness in 3D scene descriptions. Additionally, we introduce \emph{3D scenario-driven segmentation}, a benchmark that challenges models to reason about object interactions and spatial relationships beyond simple object identification. We also present Dense3D, a 3D MLLM that integrates 2D multi-view images, 3D point clouds, and textual descriptions to achieve deeper semantic understanding and more precise segmentation. We hope our work takes a step toward bridging the gap between 3D MLLMs and real-world applications, demonstrating that high-quality data and well-structured benchmarks are essential for advancing the field. 


{
    \small
    \bibliographystyle{ieeenat_fullname}
    \bibliography{main}
}

\newpage
\section*{NeurIPS Paper Checklist}

The checklist is designed to encourage best practices for responsible machine learning research, addressing issues of reproducibility, transparency, research ethics, and societal impact. Do not remove the checklist: {\bf The papers not including the checklist will be desk rejected.} The checklist should follow the references and follow the (optional) supplemental material.  The checklist does NOT count towards the page
limit. 

Please read the checklist guidelines carefully for information on how to answer these questions. For each question in the checklist:
\begin{itemize}
    \item You should answer \answerYes{}, \answerNo{}, or \answerNA{}.
    \item \answerNA{} means either that the question is Not Applicable for that particular paper or the relevant information is Not Available.
    \item Please provide a short (1–2 sentence) justification right after your answer (even for NA). 
\end{itemize}

{\bf The checklist answers are an integral part of your paper submission.} They are visible to the reviewers, area chairs, senior area chairs, and ethics reviewers. You will be asked to also include it (after eventual revisions) with the final version of your paper, and its final version will be published with the paper.

The reviewers of your paper will be asked to use the checklist as one of the factors in their evaluation. While "\answerYes{}" is generally preferable to "\answerNo{}", it is perfectly acceptable to answer "\answerNo{}" provided a proper justification is given (e.g., "error bars are not reported because it would be too computationally expensive" or "we were unable to find the license for the dataset we used"). In general, answering "\answerNo{}" or "\answerNA{}" is not grounds for rejection. While the questions are phrased in a binary way, we acknowledge that the true answer is often more nuanced, so please just use your best judgment and write a justification to elaborate. All supporting evidence can appear either in the main paper or the supplemental material, provided in appendix. If you answer \answerYes{} to a question, in the justification please point to the section(s) where related material for the question can be found.

IMPORTANT, please:
\begin{itemize}
    \item {\bf Delete this instruction block, but keep the section heading ``NeurIPS Paper Checklist"},
    \item  {\bf Keep the checklist subsection headings, questions/answers and guidelines below.}
    \item {\bf Do not modify the questions and only use the provided macros for your answers}.
\end{itemize}


\begin{enumerate}

\item {\bf Claims}
    \item[] Question: Do the main claims made in the abstract and introduction accurately reflect the paper's contributions and scope?
    \item[] Answer: \answerYes{} 
    \item[] Justification: The abstract is constructed based on the main contribution and scope of the paper.

\item {\bf Limitations}
    \item[] Question: Does the paper discuss the limitations of the work performed by the authors?
    \item[] Answer: \answerNo{}{} 
    \item[] Justification: Limitation is not discussed in the paper. 

\item {\bf Theory assumptions and proofs}
    \item[] Question: For each theoretical result, does the paper provide the full set of assumptions and a complete (and correct) proof?
    \item[] Answer: \answerNA{}{} 
    \item[] Justification:  The paper does not include any theoretical results so there's no proof attached. 

    \item {\bf Experimental result reproducibility}
    \item[] Question: Does the paper fully disclose all the information needed to reproduce the main experimental results of the paper to the extent that it affects the main claims and/or conclusions of the paper (regardless of whether the code and data are provided or not)?
    \item[] Answer: \answerYes{}{} 
    \item[] Justification: The paper explains implementation details, data annotation pipeline and the architecture in section \ref{sec:experiments} of the main paper.

\item {\bf Open access to data and code}
    \item[] Question: Does the paper provide open access to the data and code, with sufficient instructions to faithfully reproduce the main experimental results, as described in supplemental material?
    \item[] Answer: \answerNo{}{} 
    \item[] Justification: Currently we are not planning to release the code. 

\item {\bf Experimental setting/details}
    \item[] Question: Does the paper specify all the training and test details (e.g., data splits, hyperparameters, how they were chosen, type of optimizer, etc.) necessary to understand the results?
    \item[] Answer: \answerYes{} 
    \item[] Justification: Experimental Settings and detailed information about design choice are listed in the main paper.

\item {\bf Experiment statistical significance}
    \item[] Question: Does the paper report error bars suitably and correctly defined or other appropriate information about the statistical significance of the experiments?
    \item[] Answer: \answerNo{} 
    \item[] Justification: The experiment in the paper is a proof of concept and further experiment on statistical significance will be carried out in the future development. 

\item {\bf Experiments compute resources}
    \item[] Question: For each experiment, does the paper provide sufficient information on the computer resources (type of compute workers, memory, time of execution) needed to reproduce the experiments?
    \item[] Answer: \answerYes{} 
    \item[] Justification: Details of hardware compute resources is mentioned in the section \ref{sec:experiments} of the paper. 
    
\item {\bf Code of ethics}
    \item[] Question: Does the research conducted in the paper conform, in every respect, with the NeurIPS Code of Ethics \url{https://neurips.cc/public/EthicsGuidelines}?
    \item[] Answer: \answerYes{}{} 
    \item[] Justification: I have reviewed the Code of Ethics and strictly follow the rules. 

\item {\bf Broader impacts}
    \item[] Question: Does the paper discuss both potential positive societal impacts and negative societal impacts of the work performed?
    \item[] Answer: \answerYes{} 
    \item[] Justification: We have mentioned in the introduction section where the current work has great potential in the field of robotics. 
    
\item {\bf Safeguards}
    \item[] Question: Does the paper describe safeguards that have been put in place for responsible release of data or models that have a high risk for misuse (e.g., pretrained language models, image generators, or scraped datasets)?
    \item[] Answer: \answerNA{} 
    \item[] Justification: The paper does not possess such risks.

\item {\bf Licenses for existing assets}
    \item[] Question: Are the creators or original owners of assets (e.g., code, data, models), used in the paper, properly credited and are the license and terms of use explicitly mentioned and properly respected?
    \item[] Answer: \answerNA{} 
    \item[] Justification: The paper and its codebase is developed from scratch by the author without using existing assets. 

\item {\bf New assets}
    \item[] Question: Are new assets introduced in the paper well documented and is the documentation provided alongside the assets?
    \item[] Answer: \answerNA{} 
    \item[] Justification: The paper does not relase new assests so it's not applicable. 

\item {\bf Crowdsourcing and research with human subjects}
    \item[] Question: For crowdsourcing experiments and research with human subjects, does the paper include the full text of instructions given to participants and screenshots, if applicable, as well as details about compensation (if any)? 
    \item[] Answer: \answerNA{} 
    \item[] Justification: The paper does not involve crowdsourcing nor research with human subjects. 

\item {\bf Institutional review board (IRB) approvals or equivalent for research with human subjects}
    \item[] Question: Does the paper describe potential risks incurred by study participants, whether such risks were disclosed to the subjects, and whether Institutional Review Board (IRB) approvals (or an equivalent approval/review based on the requirements of your country or institution) were obtained?
    \item[] Answer: \answerNA{} 
    \item[] Justification: This paper does not involve crowdsourcing nor research with human subjects.

\item {\bf Declaration of LLM usage}
    \item[] Question: Does the paper describe the usage of LLMs if it is an important, original, or non-standard component of the core methods in this research? Note that if the LLM is used only for writing, editing, or formatting purposes and does not impact the core methodology, scientific rigorousness, or originality of the research, declaration is not required.
    \item[] Answer: \answerNA{} 
    \item[] Justification: The core method development in this research does not involve LLMs as any important, original, or non-standard components.

\end{enumerate}

\end{document}